# Technology Report : Robotic Localization and Navigation System for Visible Light Positioning and SLAM

Weipeng Guan, C. Patrick Yue
Department of Electronic and Computer Engineering, The Hong Kong University of Science and Technology, Hong Kong, SAR, China

*Abstract*—Visible light positioning (VLP) technology is a promising technique as it can provide high accuracy positioning based on the existing lighting infrastructure. However, existing approaches often require dense lighting distributions. Additionally, due to complicated indoor environments, it is still challenging to develop a robust VLP. In this work, we proposed loosely-coupled multi-sensor fusion method based on VLP and Simultaneous Localization and Mapping (SLAM), with light detection and ranging (LiDAR), odometry, and rolling shutter camera. Our method can provide accurate and robust robotics localization and navigation in LED-shortage or even outage situations. The efficacy of the proposed scheme is verified by extensive real-time experiment [1]. The results show that our proposed scheme can provide an average accuracy of 2 cm and the average computational time in low-cost embedded platforms is around 50 ms.

## I. INTRODUCTION

Precise localization is a prerequisite for many autonomous systems, such as robotics, unmanned aerial vehicle, etc. Also, the indoor localization problem is especially challenging, where localization cannot be achieved by GPS due to the satellite signal being greatly attenuated, while the traditional radio based indoor positioning technologies, such as Bluetooth, Wi-Fi, Radio-frequency Identification (RFID) and ultra-wideband (UWB), still have some disadvantages in terms of low accuracy, high latency, electromagnetic interference or high hardware cost [1].

In recent years, localization based on visible light communication (VLC), which is also referred as visible light positioning (VLP), has attracted intensive attentions as it can provide high accuracy positioning based on the existing lighting infrastructure. The modulated LED broadcasts its unique identity (ID) through switching at high frequency imperceptible to the human eye, which can be recognized by photodiodes (PD) [2]-[4] and rolling shutter effect (RSE) camera [5]-[7]. The LED lights can be mapped once for all, as they are normally fixed and not easily vulnerable to environmental changes. Hence, the "last mile problem" of localization is solved via VLP and a priori LED feature map. PD is not an ideal VLP device, since it is sensitive to the light intensity variation and the diffuse reflection of the light signal, which degrades the positioning accuracy [8]. In contrast, camera-based VLP is favored in both commerce and industry, due to the high positioning accuracy achievable by imaging geometry, and the good compatibility with user devices, such as mobile robots and smartphones. Some state-of-the-art (STOA) VLP systems can offer centimeter-level accuracy on commodity smartphone [9], [10] or mobile robot [5], [11]. Despite the promising performance of existing systems, there remain some practical challenges in VLP.

### A. Motivation

One of most urgent issue arises from the fact that VLP normally require multiple LED observations at a time for successful positioning through trilateration or triangulation. Since normal LED lights offer less usable point features, due to the lack of distinguishable appearance, e.g., one feature for a circular LED [12]. However, the number of decodable LEDs in a camera view is limited by a couple of practical factors, such as deployment density of LEDs and geometry layout, obstruction of the line-of-sight (LOS) views, limited field-of-view (FOV) of the image sensor, the ID-decoding rate cannot maintain 100%, and so on [13]. As such, the shortage or outage of LED can severely deteriorate the performance of camera-only method in reality. To address this problem, different VLP-aided inertial localization methods are proposed [5], [12]-[14], which provide pose estimation by fusing VLP measurements from the camera and the inertial measurement unit (IMU). In [13], An Extended Kalman Filter (EKF)-based loosely-coupled VLP-inertial fusion method is proposed. The presented experimental results demonstrate the functionality that the IMU helps to overcome the moments with lack of VLP coverage. However, IMU would suffer drift over time (cumulative error) or measurement noises (biases). Therefore, these VLP-aided inertial localization methods would suffer from errors during a long-term running. Based on this view, we are motivated to adopt the LiDAR scanning to compensate the accumulated error of the inertial sensor, so that the VLP-aided inertial method can achieve long-term high accurate localization. Additionally, sensors are imperfect, and their measurements are prone to errors. The measurement noise of the VLP, which is caused by fabrication error of the image sensors, is also the bottleneck of the positioning accuracy in the field of image sensor-based VLP [15]. By multi sensors fusion, we can compensate the deficiencies of stand-alone sensors and provide more reliable, more robust

---

[1] Video Demo Link: https://www.bilibili.com/video/BV1JX4y137Q7

- The robustness test when the LiDAR-SLAM is initialized incorrectly (VLP calibration for SLAM): https://www.bilibili.com/video/BV1iA411N76g/

### A. Experimental settings

We setup a room-size (around $12.0 \times 10.8$ m$^2$) test field with 4 LEDs mounted on the yellow pole with 2.7 m high (see Fig.5 (a)). We use customized LEDs (our self-designed LED[11]) and robot for experiment, as shown by Fig.5 (b) and (c). The experiments are performed on a Raspberry Pi 3B mobile robot (Turtlebot 3 Burger[12] with Quad ARM Cortex-A53 Core 1.2 GHz Broadcom BCM2837 64 bits CPU and 1 GB RAM), which runs an Ubuntu Mate 16.04 OS equipped with a robot operating system (ROS[13]). The LEDs' radiation surface has a circular shape of size 17.5cm in diameter with power rating of around 18 W. We control the RSE camera sensor (MindVision UB-300) settings by minimizing the exposure time, so as to see clear strip patterns from the modulated LEDs. The image stream is captured at around 6 Hz with a 2048 × 1536 resolution and recorded as ROS bags. The odometry and LiDAR (HLS-LFCD-LDS[14], 2D laser scanner capable of sensing 360 degrees with one-degree angular resolution) is sampled at around 24 HZ and 5 HZ, respectively. Yet for these sensors, hardware synchronization is not available. Therefore, a multithreading loose synchronization framework is used in our algorithm. We run our algorithm on a laptop computer (Intel i7-10510U CPU @1.80 GHZ, Ubuntu 18.04) using the recorded bags from the robot. The major computation is done by the computer and physical components like the LiDAR sensors, RSE-camera, odometry, etc. are interfaced through the Raspberry Pi robot, which communicates serially with the computer through publishing or subscribing the message of the related ROS packages.

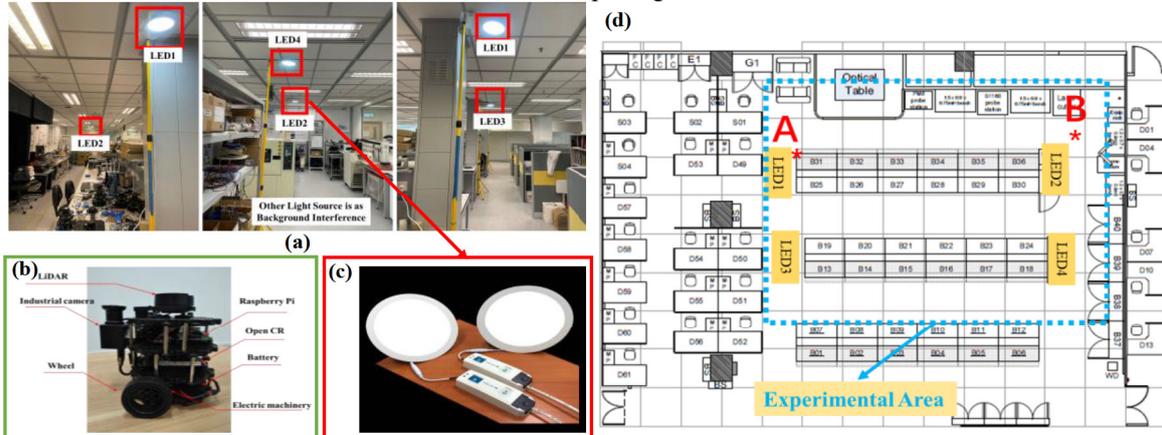

Fig.5 (a) Experimental platform; (b) The model of the robot; (c) Our self-produced LED; (d) The indoor map of our lab (A point is the origin of the prebuild LED feature map, B point is selected as a random point different from the origin).

### B. Mapping performance

Mapping is the fundamental prerequisites for localization and navigation. In this section, we firstly study the map built by Gmapping. In continuation of our previous work [13], the prebuild LED feature map is fixed and predefined along with the floorplan (blueprint) map of the building. We setup the indoor grid map of our lab (Integrated Circuit Design Center, CYT-3010, HKUST), and modified it to the occupancy grid map so that it can be visualized in the RVIZ[15]. The map made from LiDAR-SLAM is a binary map (grid map) of navigable regions and boundary walls, which are represented through white and black (boundaries and obstacles) respectively for visualization purposes. The grid map for robot localization and navigation is a representation of aspect of interest (e.g., position of landmarks, obstacles) describing the environment in which the robot operates.

The origin of the map made by traditional Gmapping is the position of the robot at the beginning of the start-up, while the origin of the LED feature map is predefined and fixed (in this paper, we set it as the Point A in Fig.5 (d)). This means that, if the robot start building Gmapping at different position, then the origin of the SLAM-map is different, which makes mismatch between the localization output from LiDAR-SLAM and VLP. With the proposed VLP-constrained Gmapping process, the VLP-map and the generated SLAM map can be aligned, so that the fusion between VLP and LiDAR-SLAM is meaningful. It is worth to mention that, in this paper, we don't evaluate the map building accuracy, since the Gmapping is a related accurate map building method [38], [39]. The comparison of grid map made through the traditional Gmapping (beginning of Point A and Point B), and the proposed VLP-constrained Gmapping (beginning of Point B) are visualized in Fig.6, where we can see that, the resulting map has similar structure.

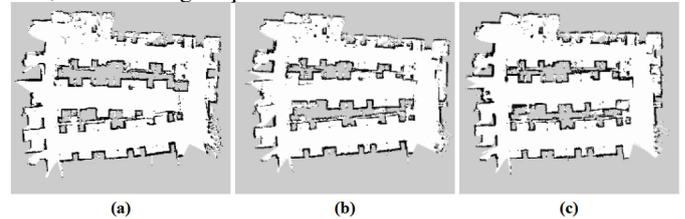

Fig.6 The resulting map through traditional Gmapping (a) start at Point A; and (b) start at Point B; (c) The resulting map through proposed VLP-constrained Gmapping start at Point B

Although the mapping performance among these three maps are similar, which fit the ground truth or the floorplan of our lab, the origin of the map is different. As can be seen in Fig.7, after the map building, we visualize the robot (stays at the same Point

---

[11] http://liphy.io/hardware/
[12] https://emanual.robotis.com/docs/en/platform/turtlebot3/overview/
[13] https://www.ros.org/
[14] http://wiki.ros.org/hls_lfcd_lds_driver
[15] http://wiki.ros.org/rviz

B with same SLO-VLP calculation result) in RVIZ. However, the position of the robot in Fig7 (b) is pretty different from Fig7 (a) and (c), which is caused by the drift of the origin between the VLP-map and the SLAM-map.

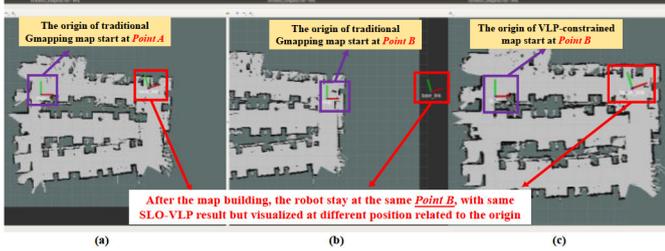

Fig.7. Using the RVIZ to visualize the position of the robot (stay at same point B; "map" coordinate is the origin of the map; "base_link" coordinate is the position of the robot through SLO-VLP calculation): (a) The SLAM-map made by traditional Gmapping start at Point A; (b) The SLAM-map made by traditional Gmapping start at Point B; (c) The SLAM-map made by proposed VLP-constrained Gmapping start at Point B

Without the VLP-constrained, the origin of the SLAM-map made from traditional Gmapping is the position of the robot at the beginning of the start-up, which may have a drift. For example, the robot starts at Point A, then the origin of the SLAM-map made from Gmapping is Point A, which is same as the origin of the LED feature map (also Point A). However, if the robot starts at Point B, then the origin of the SLAM-map made from Gmapping is Point B, which has a drift from Point A. As can be seen in Fig.8, without the VLP-constrained, the SLAM-map made from Gmapping is term the start point (Point B) of the robot as the origin of the SLAM-map. While with the VLP-constrained Gmapping, the origin of the SLAM-map can be drifted to the same point as the origin (Point A) of the pre-build LED feature map.

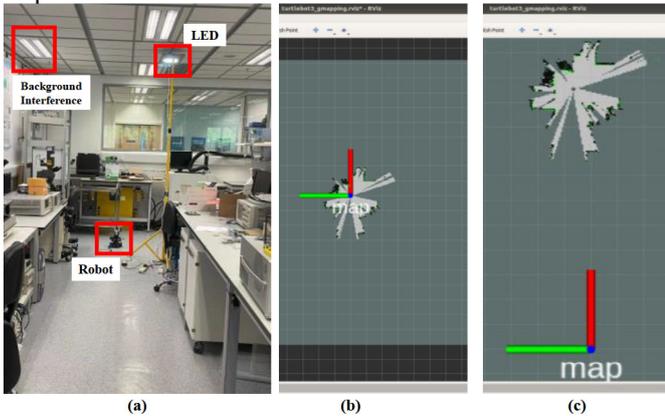

Fig.8. The grip map made at the beginning of Point B; "map" coordinate is the origin of the map. (a) The test field; (b) Traditional Gmapping; (c) The proposed VLP-constrained Gmapping

The localization output of the SLAM is based on the origin of the SLAM-map, while the localization output of the SLO-VLP is based on the origin of the LED feature map, if the origin between these two maps is not aligned, the fusion between them would be mismatched. Therefore, our VLP-constrained Gmapping can align the LiDAR scan relative to the pre-build LED feature map. The origin of the VLP-constrained Gmapping map can be shifted to the origin of the LED feature map, so that it can realize the alignment between the localization output of LiDAR-SLAM and VLP.

## C. Localization performance

### a) Positioning Accuracy

To evaluate the positioning accuracy of the proposed multi-sensor fusion for VLP-SLAM, two series of experiments were carried out. The first series are used to test the performance for motionless objects. 400 locations were randomly chosen in the experimental field to evaluate the positioning accuracy of the proposed multi-sensor fusion for VLP-SLAM. Without loss of generality, we also simultaneously calculate the positioning result from the SLO-VLP (400 samples) and the AMCL (50 samples) in the experiment. The result can be seen in Fig.9. The average positioning accuracy of the proposed multi-sensor fusion for VLP-SLAM is 1.99 cm with the maximum error of 4.97cm, which is closed to that of SLO-VLP, which is 2.59 cm with maximum error of 5.72 cm (maintain similar level as our previous work [13]). While the average accuracy of the AMCL (LiDAR-SLAM) is 7.96 cm with maximum error of 12.98 cm.

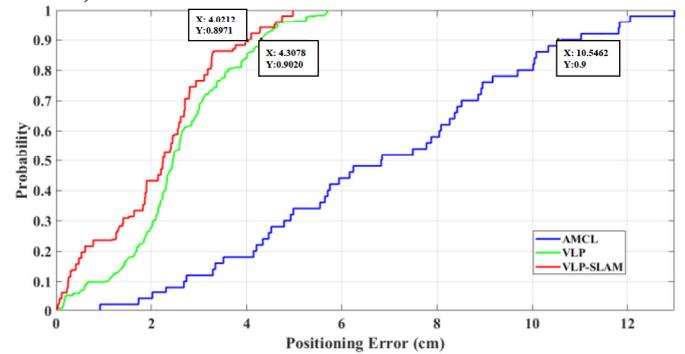

Fig.9. CDF curve of the positioning error for the VLP-SLAM fusion, SLO-VLP, and AMCL

### b) Robust and Real-time Pose Estimation

To further assess the real-time localization performance, we collected the data in trial to show the positioning performance among the SLO-VLP, AMCL, and our proposed VLP-SLAM during the movement of the robot. Fig.10 shows the twice estimated trajectories, which travel around 46 m. The video demo for these two measurements is available in[16]. Due to the limitation of our hardware, we cannot provide the motion capture system to obtain the ground-truth of a moving robot, therefore, we do not discuss the accuracy for the dynamic localization. For the proposed VLP-SLAM method, on the one hand, the SLO-VLP can provide high accuracy position, which can not only used as the pose measurement for the robot but also can be used as the pose initialization for the AMCL. On the other hand, with the AMCL and the odometry, it still can provide localization when the LED is outage (no LED coverage region). Besides, the cumulative error from the odometry can be corrected by the AMCL measurement.

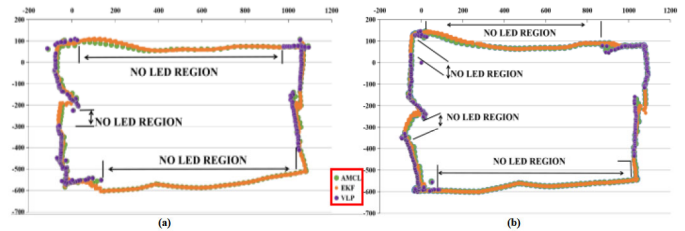

Fig.10. The trajectory estimation of our multi-sensor fusion for VLP-SLAM method with SLO-VLP and AMCL. (a) Test One and (b) Test Two for similar path.

---

[16] Our Test Demo: https://www.bilibili.com/video/BV1Zp4y1h7hF

*c) Real-time Performance*

Positioning speed is another key factor for localization systems. Especially, when the positioning terminal is moving, it can continuously receive information from the positioning system and calculate the current position in time. In this section, the computational time for position calculation of LiDAR-SLAM, SLO-VLP, and VLP-SLAM was continuously measured 300 times to calculate the average positioning time-consumption, as shown in Fig.11. The average computational time of our VLP-SLAM based on EKF is around 52ms. There are some special points of positioning time, which fluctuate from about 100 ms to 140 ms. These points are the EKF output when SLO-VLP or AMCL is available, otherwise, the mainly weight of the EKF output is calculated through the odometry sensor which is high frequency sampled. Please note that the odometry provided data at higher rates than the camera or LiDAR scanner. Even when the SLO-VLP is in-calculating (unavailable caused by calculation delay), the odometry prediction step of EKF is still used. Therefore, the derived pose estimates from VLP-SLAM can get smoothed temporally and can be sustained over a short time when LEDs are not available.

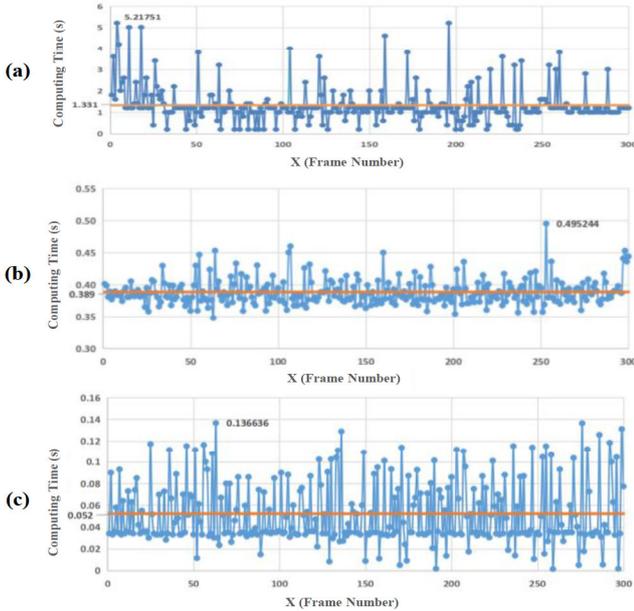

Fig.11. The measured positioning time (a) LiDAR-SLAM (AMCL); (b) SLO-VLP; (c) Multi-sensor fusion for VLP-SLAM

Note that we run the whole image process and data collection on a low-cost embedded platform Raspberry Pi 3B with a big size of captured image is 2048×1536 without any code optimization for ARM processors. After that the data (including the VLP observation, LiDAR scanning, and odometry data) are transmitted to the laptop which takes charge of multi-sensor fusion pose estimation and visualizes the result in RVIZ. The time cost here is the whole process covering both the image process and pose estimation (runtime in both laptop and the Raspberry), and also the time-cost of data transmission from robot to the laptop. Compare with the similar process platform with image size of 1640×1232 in Ref. [12], and 640×480 in Ref. [5] (Table I), the proposed multi-sensor fusion for VLP-SLAM is more efficient, and hence lightweight to be used on resource-constrained computational platform.

## D. Navigation performance

Once mapping and localization are successfully done, the navigation can be easily achieved. In this section, we embed our VLP-SLAM method into the Navigation Stack[17] in ROS to navigate the robot. While the map_server[18] and AMCL part is the optional provided node, which are replaced by our proposed occupancy grid map based on VLP-constrained Gmapping and multi-sensor fusion for VLP-SLAM localization. The distributed ROS architecture of the navigation stack based on our VLP-SLAM is shown in Fig.12.

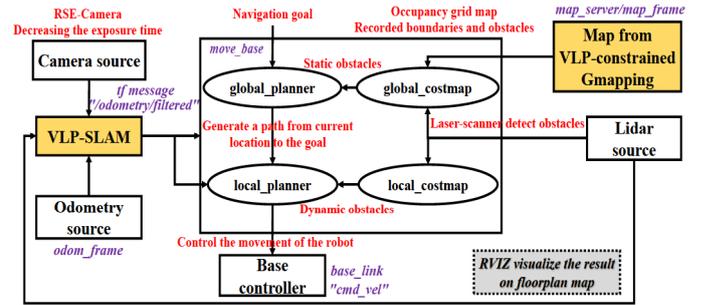

Fig.12. Visualization of the ROS computational graph for robot navigation based on our VLP-SLAM

The proposed navigation scheme uses our VLP-SLAM fusion method to estimate the pose of the robot, and the LiDAR for obstacle detection, the map data for path planning. ROS package move_base[19] is adopted to accomplish autonomous navigation. This package uses the localization information from our multi-sensor fusion and provides move-commands to the mobile base of robot to move safely in the environment without colliding with obstacle. The framework also maintains two costmap[20] each for the local and global planner. The global planner is based on A Star search algorithm, while the local planner used the dynamic window approach [40]. Once a navigation goal is received, the global path planning algorithm takes charge of generating an appropriate path from the start pose to the target pose, while the local path planning is responsible for generating velocity commands based on the pose of the robot which is calculated through the our VLP-SLAM fusion. During the automatically navigation, the dynamic obstacles can be detected through the 2D LiDAR. Through the update of the local costmap to realize the dynamic obstacle avoidance. The accurate VLP pose estimation contributes to pose initialization for global path planning, while the real-time and robust VLP-SLAM pose estimation helps to local path planning. For the details about the navigation test, we strongly recommend the readers to our video demo:

- https://www.bilibili.com/video/BV1JX4y137Q7

## E. Discussion

In this section, we compare the performance of our VLP-SLAM with the SOTA works in the field of VLP in Table I. The average positioning accuracy, computing time, and the density of the LEDs in the related experimental platform are also reported objectively. It is obvious that the density of the LEDs in our VLP-SLAM system achieve state-of-the-art, with less latency (52ms) and high accuracy (2.0 cm). The proposed approach can work with sparse lighting distributions, which greatly increases the coverage of the effective positioning area. For a large area of (12.0×10.8×2.7 $m^3$) only 4 LEDs are required to ensure high accuracy positioning, while the other works cannot meet this high requirement. Compared to other sensor fusion approaches, our method perform favorably. Especially for the VLP-aided IMU method [5], [12]-[14], which suffers from bias or drift for long-term running. Our work can cope with the cumulative error through the LiDAR scan matching. Through the multi-sensor fusion, especially the LiDAR scanning compensation, the orientation angle from the odometry can perform well with considerable drifts after long-term experiment. So that the single LED VLP calculation, which is based on the orientation angle from the odometry, can obtain low-drift motion estimation. What's more, different harsh situations, such as without LED, background light interference, handover, etc., are included during our test. Therefore, the proposed multi-sensor fusion for VLP-SLAM can achieve well balance among accuracy, real-time performance, and robustness. The multi-sensor fusion can compensate the deficiencies of stand-alone sensors and provide more reliable, more robust positioning and navigation.

---

[17] http://wiki.ros.org/navigation/Tutorials/RobotSetup
[18] http://wiki.ros.org/map_server
[19] http://wiki.ros.org/move_base
[20] http://wiki.ros.org/costmap_2d

The LiDAR sensor make great contributions for the reliable localization and navigation. However, for LiDAR-SLAM, it is difficult to estimate the pose in the large space or long corridors without too much variety of observation. Since similar geometric regions in these environment may mislead the convergence of particles in AMCL. For example, the geometry of our indoor environment looks likely in region B19-B24 and B07-B12 (Fig.5d). When the robot is not under the VLP coverage area, the wrong initialization for LiDAR-SLAM would make the robot getting lost since the geometric region of these two parts is quite similar. However, when the VLP is available, the position of the robot can be calibrated to the right position. The evaluation about the VLP calibration for LiDAR-SLAM can be seen in the video demo[21].

Besides, initialization is one of the crucial problems that needs to be solved in order to achieve fully autonomous navigation. For LiDAR-SLAM, the AMCL needs to initialize its filters, this is generally done with an initial pose message given by the user. In our VLP-SLAM method, the VLP can provide position initialization for the AMCL, while the AMCL can provide reliable orientation estimation for the SLO-VLP. Based on this, the robot is no longer need to head along fixed direction (e.g. X-axes) before running. Even if the manually initial pose (position and orientation) of the robot is not accurate, the LiDAR-SLAM can adjust the orientation of the robot, while the SLO-VLP can provide higher accuracy position estimation with the more correct orientation from the LiDAR. Meanwhile, the high position estimation from SLO-VLP can update overall pose estimation for the EKF, which can also act on the AMCL to improve the measurement from the AMCL. The interaction between VLP and LiDAR-SLAM provide reliable and accuracy pose estimation. The details about this can be visualized through the video demo[22].

Admittedly, the proposed system still suffers a few limitations. The high accuracy VLP calculation is based on the accurate prebuild LED feature map. In other words, the actual installation position of the LED lamps is very important for the positioning accuracy. Especially, the automatic construction for the LED beacon in large scale scenarios. In future work, we would like to work on the tightly coupled VLP and SLAM through incorporating the LED markers into the maps and refining the SLAM map model simultaneously. On the one hand, it can generate the LED feature map with the occupancy grid map. On the other hand, the observation of external LED landmarks can be useful to deduce the trajectory drift and possibly correct the LiDAR-SLAM.

Table I. Performance comparison with the SOTA

| Method | Average Accuracy (cm) | Time Cost (ms) | Require LEDs (at least) | Hardware | Dimension | Receiver Type | Density LEDs-Coverage (L× W × H)$m^3$ |
|---|---|---|---|---|---|---|---|
| R. [41] | 14.0 | N/A | 3 | STM32F103+computer | 2D | PD+IMU | 7 -(1.5×1.5×2.5) |
| R. [1] | 3.9 | 44 | 2 | Raspberry+computer | 3D+yaw | Camera | 9 -(1.8×1.8×2.0) |
| R. [11] | 0.8 | 400 | 2 | Raspberry+computer | 3D+yaw | Camera | 4 -(1.0×1.0×1.5) |
| R. [26] | 17.5 | N/A | 1* | Smartphone | 2D+3D | Camera | 1 -(3.0×3.0×2.0) |
| R. [42] | 3.2 | 43 | 1* | computer | 3D+yaw | Camera | 1 -(0.8×0.8×2.0) |
| R. [43] | 2.3 | 60 | 1* | computer | 2D+yaw | Camera | 1 -(1.8×1.8×3.0) |
| R. [27] | 5.5 | N/A | 1 | Smartphone | 6D | Camera+IMU | 1 -(2.7×1.8×1.5) |
| R. [8] | 13.4 | N/A | 1 | Smartphone | 3D+yaw | Camera+PDR | 1 -(1.0×1.0×2.6) |
| R. [14] | 5.0 | N/A | 1 | Raspberry+computer | 3D+yaw | Camera+IMU | 23 -(5.0×4.0×2.3) |
| R. [12] | 5.0 | 50 | 1 | Raspberry+computer | 3D+yaw | Camera+IMU | 25 -(5.0×4.0×2.3) |
| R. [5] | 3.0 | 78# | 1 | Raspberry+computer | 2D+yaw | Camera+encoder+gyroscope | 4 -(2.0×2.0×1.5) |
| R. [5] | 3.0 | 131.8# | 1 | Raspberry+computer | 2D+yaw | Camera+encoder+gyroscope | 4 -(2.0×2.0×1.5) |
| R. [5] | 2.2 | 81.5# | 1 | Raspberry+computer | 2D+yaw | Camera+encoder+gyroscope | 4 -(2.0×2.0×1.5) |
| R. [13] | 2.0 | 195 | 1 | Raspberry+computer | 3D+yaw | Camera+odom | 3 -(7.0×3.8×2.7) |
| R. [13] | 2.1 | 33 | 1 or 0 | Raspberry+computer | 3D+yaw | Camera+odom+IMU | 3 -(9.1×4.0×2.7) |
| *Ours* | 2.0 | 52 | 1 or 0 | Raspberry+computer | 3D+yaw | Camera+odom+LiDAR | 4 -(12.0×10.8×2.7) |

"*" means that the LED lamp with beacon marker
"#" means that the position estimation was performed in post-processing, not real-time.

## V. CONCLUSION

In this paper, we pursue reliable, real-time, and accurate pose estimation for mobile robot through the multi-sensor fusion method based on EKF for VLP-SLAM. We relaxed the assumption on the minimum number of concurrently observable LEDs from three to zero for the RSE-camera based VLP system through the fusion with odometry and LiDAR sensor. While the cumulative error of the odometry is compensated through the LiDAR scanning. We also embed our multi-sensor fusion for VLP-SLAM into the navigation stack of ROS, owing to the high accuracy performance of our multi-sensor fusion scheme, the robot can move steadily according to the planned path to reach the designated location. The result shows that the our method has strong robustness with low latency (52ms) and high accuracy (2.0 cm), which can keep well balance among accuracy, real-time ability and coverage. In future work, we will deeply study the tightly coupled VLP and SLAM for incorporating the LED markers into the maps, which could construct the global map and optimize poses with sufficient features together with a method to model and reduce data uncertainty.

---

[21] VLP calibration for SLAM:
https://www.bilibili.com/video/BV1iA411N76g/

[22] Pose initialization for VLP-SLAM:
https://www.bilibili.com/video/BV1VX4y1G7wN

## VI. REFERENCES

[1] P. Lin *et al*, "Real-time visible light positioning supporting fast moving speed," *Optics Express,* vol. 28, *(10),* pp. 14503-14510, 2020.
[2] N. Huang *et al*, "Design and Demonstration of Robust Visible Light Positioning Based on Received Signal Strength," *J. Lightwave Technol.,* vol. 38, *(20),* pp. 5695-5707, 2020.
[3] Y. Cai *et al*, "Indoor high precision three-dimensional positioning system based on visible light communication using particle swarm optimization," *IEEE Photonics Journal,* vol. 9, *(6),* pp. 1-20, 2017.
[4] B. Chen *et al*, "Performance comparison and analysis on different optimization models for high-precision three-dimensional visible light positioning," *Optical Engineering,* vol. 57, *(12),* pp. 125101, 2018.
[5] R. Amsters *et al*, "Visible Light Positioning Using Bayesian Filters," *J. Lightwave Technol.,* vol. 38, *(21),* pp. 5925-5936, 2020.
[6] Y. Yang, J. Hao and J. Luo, "CeilingTalk: Lightweight indoor broadcast through LED-camera communication," *IEEE Transactions on Mobile Computing,* vol. 16, *(12),* pp. 3308-3319, 2017.
[7] Y. Kuo *et al*, "Luxapose: Indoor positioning with mobile phones and visible light," in *Proceedings of the 20th Annual International Conference on Mobile Computing and Networking,* 2014, .
[8] H. Huang *et al*, "Hybrid indoor localization scheme with image sensor-based visible light positioning and pedestrian dead reckoning," *Appl. Opt.,* vol. 58, *(12),* pp. 3214-3221, 2019.
[9] A. Jovicic, "Qualcomm Lumicast: A high accuracy indoor positioning system based on visible light communication," *White Paper, Qualcomm, Apr,* 2016.
[10] J. Fang *et al*, "High-speed indoor navigation system based on visible light and mobile phone," *IEEE Photonics Journal,* vol. 9, *(2),* pp. 1-11, 2017.
[11] W. Guan *et al*, "High-Accuracy Robot Indoor Localization Scheme based on Robot Operating System using Visible Light Positioning," *IEEE Photonics Journal,* vol. 12, *(2),* pp. 1-16, 2020.
[12] Q. Liang and M. Liu, "A tightly coupled VLC-inertial localization system by EKF," *IEEE Robotics and Automation Letters,* vol. 5, *(2),* pp. 3129-3136, 2020.
[13] W. Guan *et al*, "Robust Robotic Localization using Visible Light Positioning and Inertial Fusion," *IEEE Sensors Journal,* pp. 1, 2021. . DOI: 10.1109/JSEN.2021.3053342.
[14] Q. Liang, J. Lin and M. Liu, "Towards robust visible light positioning under LED shortage by visual-inertial fusion," in *2019 International Conference on Indoor Positioning and Indoor Navigation (IPIN),* 2019, pp. 1-8.
[15] S. Chen and W. Guan, "High Accuracy VLP based on Image Sensor using Error Calibration Method," *arXiv Preprint arXiv:2010.00529,* 2020.
[16] T. Caselitz *et al*, "Monocular camera localization in 3d lidar maps," in *2016 IEEE/RSJ International Conference on Intelligent Robots and Systems (IROS),* 2016, .
[17] S. Chan, P. Wu and L. Fu, "Robust 2D indoor localization through laser SLAM and visual SLAM fusion," in *2018 IEEE International Conference on Systems, Man, and Cybernetics (SMC),* 2018, .
[18] V. Nguyen, A. Harati and R. Siegwart, "A lightweight SLAM algorithm using orthogonal planes for indoor mobile robotics," in *2007 IEEE/RSJ International Conference on Intelligent Robots and Systems,* 2007, .
[19] P. Hu *et al*, "High speed LED-to-Camera communication using color shift keying with flicker mitigation," *IEEE Transactions on Mobile Computing,* 2019.
[20] C. Xie et al, "The LED-ID detection and recognition method based on visible light positioning using proximity method," IEEE Photonics Journal, vol. 10, (2), pp. 1-16, 2018.
[21] W. Guan *et al*, "Performance analysis and enhancement for visible light communication using CMOS sensors," *Opt. Commun.,* vol. 410, pp. 531-551, 2018.
[22] Y. Xiao et al, "The Optical Bar Code Detection Method Based on Optical Camera Communication Using Discrete Fourier Transform," IEEE Access, vol. 8, pp. 123238-123252, 2020.
[23] W. Guan *et al*, "High-precision indoor positioning algorithm based on visible light communication using complementary metal–oxide–semiconductor image sensor," *Optical Engineering,* vol. 58, *(2),* pp. 024101, 2019.
[24] W. Guan *et al*, "High precision indoor visible light positioning algorithm based on double LEDs using CMOS image sensor," *Applied Sciences,* vol. 9, *(6),* pp. 1238, 2019.
[25] H. Lee *et al*, "Rollinglight: Enabling line-of-sight light-to-camera communications," in *Proceedings of the 13th Annual International Conference on Mobile Systems, Applications, and Services,* 2015, .
[26] R. Zhang *et al*, "A single LED positioning system based on circle projection," *IEEE Photonics Journal,* vol. 9, *(4),* pp. 1-9, 2017.
[27] H. Cheng *et al*, "A Single LED Visible Light Positioning System Based on Geometric Features and CMOS Camera," *IEEE Photonics Technology Letters,* vol. 32, *(17),* pp. 1097-1100, 2020.
[28] C. Cadena *et al*, "Past, present, and future of simultaneous localization and mapping: Toward the robust-perception age," *IEEE Transactions on Robotics,* vol. 32, *(6),* pp. 1309-1332, 2016.
[29] H. Ye, Y. Chen and M. Liu, "Tightly coupled 3d lidar inertial odometry and mapping," in *2019 International Conference on Robotics and Automation (ICRA),* 2019, pp. 3144-3150.
[30] W. Hess *et al*, "Real-time loop closure in 2D LIDAR SLAM," in *2016 IEEE International Conference on Robotics and Automation (ICRA),* 2016, .
[31] B. Steux and O. El Hamzaoui, "tinySLAM: A SLAM algorithm in less than 200 lines C-language program," in *2010 11th International Conference on Control Automation Robotics & Vision,* 2010, .
[32] M. Montemerlo *et al*, "FastSLAM: A factored solution to the simultaneous localization and mapping problem," *Aaai/Iaai,* vol. 593598, 2002.
[33] S. Kohlbrecher *et al*, "Hector open source modules for autonomous mapping and navigation with rescue robots," in *Robot Soccer World Cup,* 2013, .
[34] G. Grisetti, C. Stachniss and W. Burgard, "Improved techniques for grid mapping with rao-blackwellized particle filters," *IEEE Transactions on Robotics,* vol. 23, *(1),* pp. 34-46, 2007.
[35] G. Grisettiyz, C. Stachniss and W. Burgard, "Improving grid-based slam with rao-blackwellized particle filters by adaptive proposals and selective resampling," in *Proceedings of the 2005 IEEE International Conference on Robotics and Automation,* 2005, .
[36] W. Guan *et al*, "High-speed robust dynamic positioning and tracking method based on visual visible light communication using optical flow detection and Bayesian forecast," *IEEE Photonics Journal,* vol. 10, *(3),* pp. 1-22, 2018.
[37] B. Hussain, C. Qiu and C. P. Yue, "A universal VLC modulator for retrofitting LED lighting and signage," in *2019 IEEE 8th Global Conference on Consumer Electronics (GCCE),* 2019, .
[38] J. M. Santos, D. Portugal and R. P. Rocha, "An evaluation of 2D SLAM techniques available in robot operating system," in *2013 IEEE International Symposium on Safety, Security, and Rescue Robotics (SSRR),* 2013, .
[39] K. Krinkin *et al*, "Evaluation of modern laser based indoor slam algorithms," in *2018 22nd Conference of Open Innovations Association (FRUCT),* 2018, .
[40] E. Marder-Eppstein *et al*, "The office marathon: Robust navigation in an indoor office environment," in *2010 IEEE International Conference on Robotics and Automation,* 2010, .
[41] Z. Li *et al*, "Fusion of visible light indoor positioning and inertial navigation based on particle filter," *IEEE Photonics Journal,* vol. 9, *(5),* pp. 1-13, 2017.
[42] W. Guan *et al*, "A novel three-dimensional indoor localization algorithm based on visual visible light communication using single LED," in *2018 IEEE International Conference on Automation, Electronics and Electrical Engineering (AUTEEE),* 2018, .
[43] H. Li *et al*, "A Fast and High-accuracy Real-time Visible Light Positioning System based on Single LED Lamp with a Beacon," *Jphot,* pp. 1, 2020. Available: https://ieeexplore.ieee.org/document/9233945. DOI: 10.1109/JPHOT.2020.3032448.